\documentclass[runningheads]{llncs}
\usepackage[T1]{fontenc}
\usepackage{amsmath}
\usepackage{amssymb}
\usepackage{booktabs}
\usepackage{caption}
\usepackage{graphicx}
\begin{document}
\title{RadarSeq: A Temporal Vision Framework for User Churn Prediction via Radar Chart Sequences}

\author{Sina Najafi\inst{1} \and
M. Hadi Sepanj\inst{2}  \and
Fahimeh Jafari\inst{1}} 

\authorrunning{S. Najafi et al.}
\institute{
University of East London, London, UK \\
\email{u2735128@uel.ac.uk, F.Jafari@uel.ac.uk}
\and
Vision Group, Systems Design Engineering, University of Waterloo, Waterloo, Canada \\
\email{mhsepanj@uwaterloo.ca}
}
% \author{Anonymous Submission}

\maketitle              % typeset the header of the contribution

\begin{abstract}
Predicting user churn in non-subscription gig platforms, where disengagement is implicit, poses unique challenges due to the absence of explicit labels and the dynamic nature of user behavior. Existing methods often rely on aggregated snapshots or static visual representations, which obscure temporal cues critical for early detection. In this work, we propose a temporally-aware computer vision framework that models user behavioral patterns as a sequence of radar chart images, each encoding day-level behavioral features. By integrating a pretrained CNN encoder with a bidirectional LSTM, our architecture captures both spatial and temporal patterns underlying churn behavior. Extensive experiments on a large real-world dataset demonstrate that our method outperforms classical models and ViT-based radar chart baselines, yielding gains of +17.7 in F1 score, +29.4 in precision, and +16.1 in AUC, along with improved interpretability. The framework’s modular design, explainability tools, and efficient deployment characteristics make it suitable for large-scale churn modeling in dynamic gig-economy platforms.

\end{abstract}

\section{Introduction}\label{sec:intro}

Churn prediction is a long-standing problem in the fields of marketing analytics~\cite{blattberg2009customer,fader2010customer,gupta2008research}, operations~\cite{lu2021review,donohue2020behavioral,paiva2024developing}, and customer lifecycle modeling~\cite{verbraken2012novel,neslin2006challenges,mozer2002evidence}. The ability to identify disengaging users early enables companies to deploy personalized retention strategies and mitigate revenue loss~\cite{east2001customer,reichheld1990zero}. In traditional subscription-based domains such as telecommunications, churn is clearly defined and easily labeled~\cite{al_janabi2018intelligent,imani2024customer,krishnan2021customer}. However, in non-subscription-based services~\cite{burez2009handling,gattermann2022proactive,amsal2021systematic} such as e-commerce~\cite{yang2021rf,matuszelanski2022customer} or gig platforms~\cite{munoz2022customer,hall2018analysis}, churn is often latent and can only be inferred retrospectively from behavioral inactivity~\cite{de2022predicting,torfs2019vitro,bernat2019modelling}. This makes early detection both more difficult and more valuable.

On-demand delivery platforms~\cite{mukhopadhyay2020your,sun2024dinner,wang2025structural} experience notably high churn rates, often exceeding 40\% annually. For instance, Uber has reported that only 4\% of newly recruited drivers remain active after one year~\cite{oblander2021has,benner2020delivering}, while similar trends are documented across platforms like DoorDash and Deliveroo~\cite{heing2020gigecon,ganapathy2025decoding}. This churn epidemic necessitates scalable predictive solutions. Based on operational insights, a courier inactive for 45 consecutive days is typically considered lost to competitors or other employment—thus providing a robust, real-world validated definition of churn~\cite{al2023analyzing,guo2024seeking}.

From an economic perspective, courier retention is significantly more cost-effective than replacement. Industry estimates indicate that recruiting and onboarding a new courier can cost between \$8,000 and \$10,000 per individual~\cite{munoz2022customer}. These figures account for direct costs such as training and lost productivity, and often exclude secondary impacts like service disruptions and safety risks associated with inexperienced workers. In contrast, proactive churn prevention—e.g., offering targeted incentives or scheduling flexibility to at-risk couriers—can often retain them at a fraction of the cost. This return on investment is amplified when churn prediction models are used to focus retention efforts only where needed~\cite{katyayan2025optimizing}. By reducing turnover, platforms not only lower operational expenses but also benefit from a more stable, experienced, and efficient workforce, directly enhancing long-term profitability and scalability~\cite{andersson2024developing}.

A common strategy for churn modeling involves aggregating behavioral features—such as order frequency, transaction amounts, and recency—into static summary statistics~\cite{hadden2007computer,buckinx2005customer,chen2015predicting}, a simplification that our method explicitly avoids by modeling temporal behavior directly. Classical machine learning methods (e.g., logistic regression~\cite{khodabandehlou2017comparison,vafeiadis2015comparison}, random forests~\cite{idris2012churn,sharma2022enhancing}, and gradient boosting models~\cite{sung2017evaluating,bogaert2023ensemble}) have shown strong performance when trained on such feature sets~\cite{ahmed2024comparative,chang2024prediction}. However, these approaches discard fine-grained temporal patterns in user behavior, potentially missing early indicators of disengagement~\cite{zhang2015predicting}.

To overcome the limitations of tabular modeling, recent research has explored the visual encoding of behavior logs. Coolwijk et al.~\cite{coolwijk2025vision} introduced a Vision Transformer~\cite{dosovitskiy2021thomas} (ViT)-based pipeline where aggregated customer features are converted into radar chart images, enabling the use of pre-trained image classification models for churn detection. While effective, this approach relies on a single static image per user, abstracting away the evolution of user behavior over time.

In this work, we argue that temporal structure is essential for early and accurate churn prediction. Rather than summarizing behavior, we propose to model it as a time-indexed visual sequence. Specifically, we transform each user's daily transaction profile into a radar chart image and model the sequence of images using a convolutional-recurrent architecture~\cite{donahue2015long,srivastava2015unsupervised,yue2015beyond}. This preserves behavioral dynamics and enables temporal pattern recognition across multiple scales. Radar charts preserve spatial relationships between multi-dimensional data points, maintaining contextual dependencies and enabling visual feature learning via CNNs~\cite{jastrzebska2020lagged}. Unlike heatmaps or recurrence plots, radar charts offer consistent geometric encoding of features, enhancing interpretability and preserving domain semantics across time steps. Prior research supports their value in temporal feature retention and interpretability~\cite{dong2023retention,zhang2022explainable,sacha2016visual}. Our method represents the first known attempt to use radar chart imagery combined with CNN+LSTM~\cite{ullah2017action} architectures for churn prediction, thus capturing complex spatio-temporal behavioral patterns. Our contributions are as follows:
\begin{itemize}
    \item We introduce a novel vision-based framework for churn prediction that preserves the full temporal granularity of user behavior by generating daily radar chart images.
    \item We propose a hybrid CNN+LSTM architecture that learns both spatial patterns within daily behavior encodings and temporal dependencies across time.
    \item We conduct extensive experiments comparing our approach against traditional ML models and the ViT-based radar chart baseline. Our method achieves state-of-the-art performance on multiple metrics, including AUC~\cite{xie2020motion}, precision~\cite{tatsunami2022sequencer}, and F1-score~\cite{yang2018making}.
\end{itemize}

By combining visual representation learning~\cite{zhang2018network,sepanj2025self} with temporal modeling~\cite{cheng2024videollama,sepanj2025uncertainty,ding2024joint}, our method provides a generalizable and interpretable framework for user behavior modeling in domains where churn is implicit and dynamically emerging, and these visual temporal sequences allow the application of spatial-temporal deep learning architectures.

Accurately predicting courier churn has direct economic implications for on-demand delivery platforms. Studies indicate that retaining existing couriers is up to five times more cost-effective than recruiting and onboarding new workers~\cite{blattberg2008database,dai2021cost}. Effective churn prediction reduces recruitment expenditure, minimizes downtime in delivery operations, and enhances user satisfaction through consistent courier availability~\cite{silverman2022gig}. At scale, a modest 5--10\% improvement in retention can lead to substantial savings in operational costs and platform incentives~\cite{dong2023retention}.

\section{Background}

User churn prediction remains a critical problem in e-commerce and gig platforms, where users do not explicitly unsubscribe but instead silently disengage~\cite{khodabandehlou2017comparison}. Initial studies in churn modeling relied on structured tabular features~\cite{matuszelanski2022customer} and classical machine learning models~\cite{sina2022model} such as logistic regression~\cite{granov2021customer}, decision trees~\cite{granov2021customer}, support vector machines (SVMs)~\cite{khodabandehlou2017comparison}, and ensemble methods including random forests~\cite{granov2021customer}, XGBoost~\cite{panimalar2023review}, and CatBoost~\cite{huang2012customer,buckinx2005customer,imani2024customer}. While interpretable and computationally efficient, these models typically depend on temporal aggregation—often through Recency-Frequency-Monetary (RFM) variables—which discards short-term behavioral fluctuations crucial for early churn detection~\cite{aleksandrova2018application}.

To better capture user dynamics, deep learning models such as Long Short-Term Memory (LSTM)~\cite{mena2024exploiting} networks and Gated Recurrent Units (GRUs)~\cite{zhang2023brief} have been explored for sequence modeling in churn prediction~\cite{krishnan2021customer,zhang2023comparative,liu2024deep}. Transformers have further extended this capability via parallelized attention mechanisms~\cite{kougioumtzidis2025mobile} that model global dependencies~\cite{vaswani2017attention,ren2021predicting,zahin2023multi}. Despite their success, these models are rarely combined with visual encoding techniques.

Recent efforts have introduced vision-based methods for behavioral modeling. Wang et al.~\cite{wang2015imaging} and Li et al.~\cite{li2023time} proposed converting time-series data into image representations—such as Gramian Angular Fields~\cite{wang2015imaging} or radar plots—enabling the application of convolutional neural networks (CNNs) to discover spatial patterns. Coolwijk et al.~\cite{coolwijk2025vision} advanced this idea by transforming RFM features into radar chart images and applying a Vision Transformer (ViT) pretrained on ImageNet~\cite{coolwijk2025vision,dosovitskiy2020image} for churn classification. Their pipeline, inspired by ChurnViT~\cite{rabbah2023new}, clusters customers via k-means~\cite{lloyd1982least} to generate pseudo-churn labels. Although promising in recall and AUC, this approach presents several drawbacks:
\begin{itemize}
    \item \textbf{Static Representation:} Each customer is encoded into a single radar chart summarizing their entire behavioral history, eliminating temporal continuity and masking changes in behavior over time.
    \item \textbf{No Sequential Modeling:} ViTs classify each image independently, making them less suited for temporal modeling tasks such as churn prediction.
    \item \textbf{Interpretability Challenges:} Radar charts with non-aligned features and attention-based decisions obscure transparency, especially when no temporal context is preserved.
\end{itemize}

\vspace{-.22cm}
In contrast, temporal models like LSTMs and temporal CNNs are well-suited for capturing sequential patterns in behavior~\cite{wang2024risk,sheil2018predicting}. However, their integration with visual encodings remains underexplored. Most prior work either uses raw tabular features or aggregates transactional logs into coarse temporal bins before classification~\cite{ahn2021modeling}, which compromises the temporal resolution of the user history.

Our proposed method addresses these limitations by transforming each day of the user behavior pattern into a radar chart image, preserving both visual and temporal information. These daily radar charts form an image sequence input to a hybrid CNN+LSTM model: the CNN encodes spatial feature patterns, and the LSTM captures temporal evolution. This design retains the interpretability of image-based models while enabling trajectory-level modeling of user behavior.
\vspace{-.6cm}
\begin{table}[ht]
\centering
\caption{Comparison of churn prediction methods across key design dimensions}
\label{tab:method-comparison}
\begin{tabular}{lcccc}
\toprule
\textbf{Model} & \textbf{Temporal} & \textbf{Visual} & \textbf{Input Type} & \textbf{Supervision} \\
\midrule
CatBoost & No & No & Tabular & Explicit \\
ChurnViT & No & No & Radar Img & Pseudo \\
Proposed CNN+LSTM & Yes & Yes & Radar Seq & Explicit \\
\bottomrule
\end{tabular}
\end{table}
\vspace{-.6cm}
In doing so, our method generalizes the radar-ViT framework to a temporal regime. This integration of visual encoding and recurrent architectures enables early and accurate churn prediction, as demonstrated by our empirical improvements in F1-score, precision, and AUC. Our framework, therefore, surpasses both traditional ML and static vision-based approaches in performance while offering greater interpretability through spatial-temporal decomposition~\cite{yang2022cnn}.

\section{Methodology}

We propose a temporally-aware, image-based churn prediction framework that processes each biker's behavioral timeline as a sequence of daily radar chart images. The core idea is to preserve behavioral dynamics over time while enabling visual representation learning through a hybrid CNN+LSTM architecture. Unlike prior approaches that compress user history into a single image for Vision Transformer classification \cite{coolwijk2025vision}, our method retains temporal structure and enables trajectory-level modeling.

\subsection{Data Representation}

Let each biker $c \in \mathcal{C}$ have a historical timeline of daily activities represented as a sequence of feature vectors:
\[
\mathbf{x}_c = \left\{ \mathbf{x}_c^{(1)}, \mathbf{x}_c^{(2)}, \dots, \mathbf{x}_c^{(T)} \right\}, \quad \mathbf{x}_c^{(t)} \in \mathbb{R}^{d}
\]
where $d$ is the number of behavioral features (e.g., earnings, trip count, ride duration, etc.) and $T$ is the number of days in the sliding window. Each $\mathbf{x}_c^{(t)}$ is transformed into a radar chart image $\mathbf{I}_c^{(t)} \in \mathbb{R}^{H \times W}$, using a polar plot rendering function $\mathcal{R}: \mathbb{R}^d \to \mathbb{R}^{H \times W}$. We choose $(H, W) = (32, 32)$ for efficiency.

\begin{figure}[ht]
\centering
\begin{tabular}{cc}
    \includegraphics[width=0.4\linewidth]{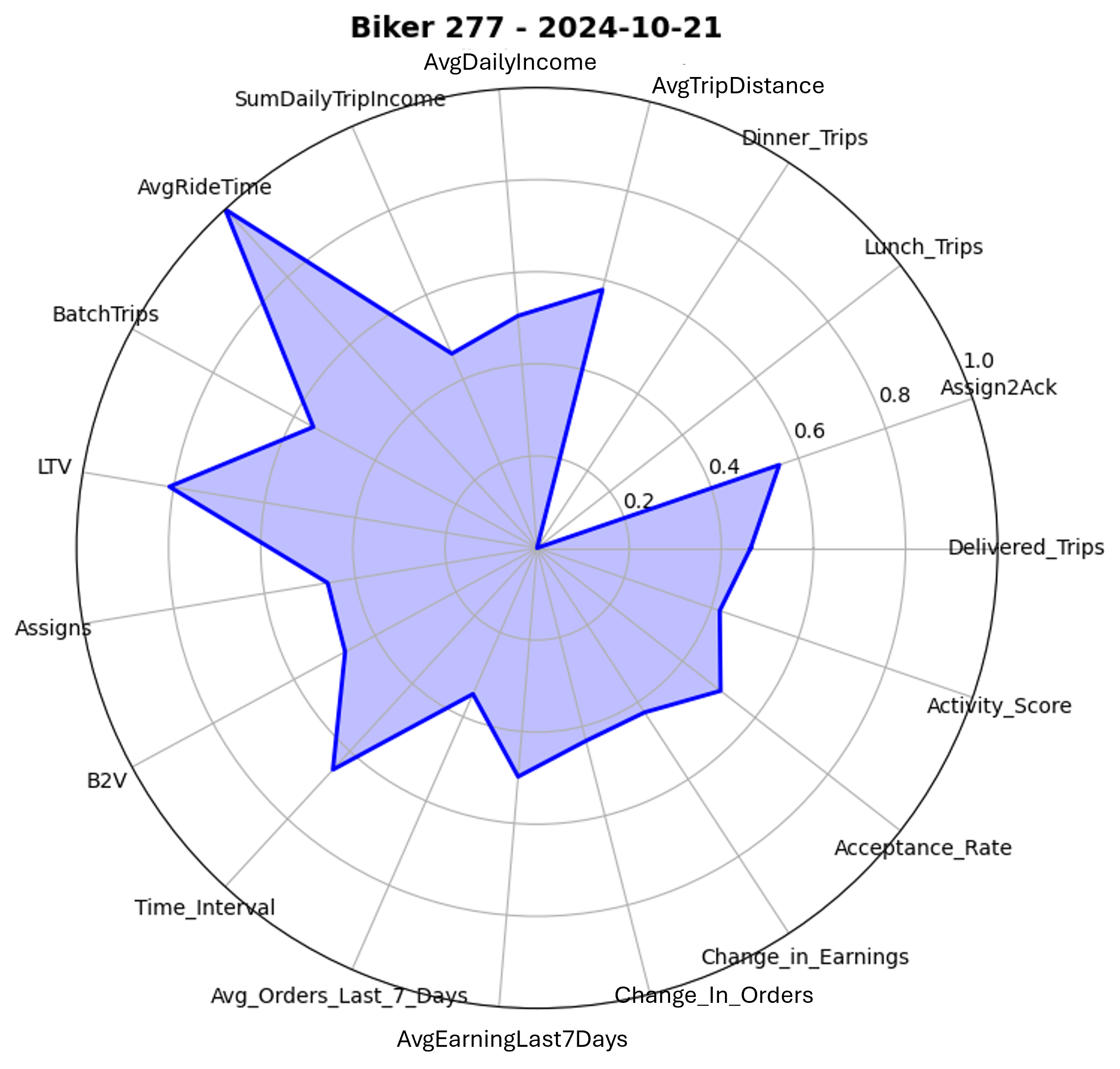} &
    \includegraphics[width=0.4\linewidth]{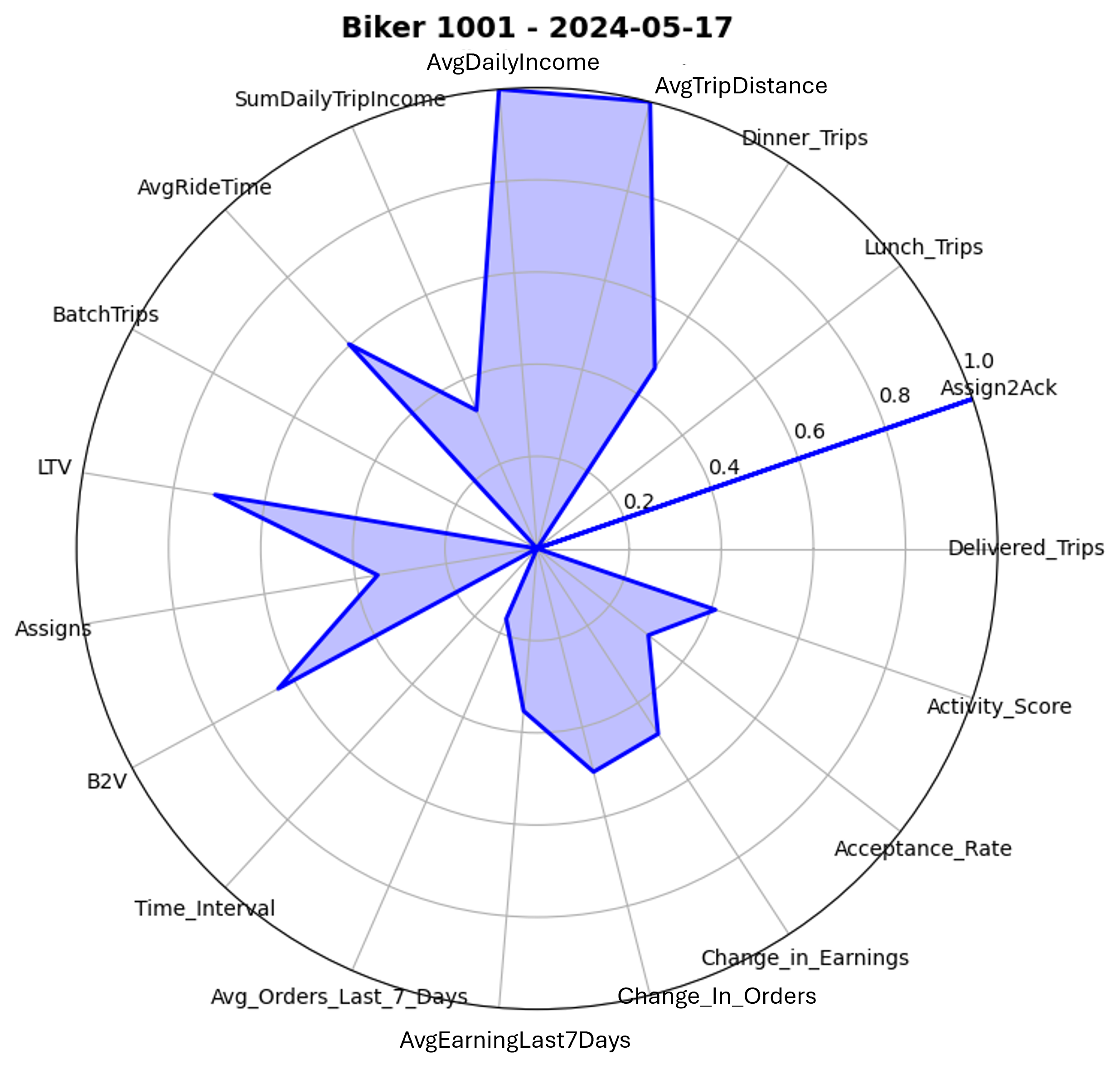} \\
    \textbf{(a)} A day of biker \emph{a} behavior & \textbf{(b)} A day of biker \emph{b} behavior  \\
\end{tabular}
\caption{
Examples of radar chart images representing daily courier behavior. Each chart encodes multivariate features such as earnings, trip count, and idle time. Temporal differences in chart shapes reflect changes in behavioral patterns over time, which our CNN+LSTM model captures to predict churn.
}
\label{fig:radar_example}
\end{figure}

The resulting input is a spatiotemporal image sequence:
\[
\mathbf{I}_c = \left\{ \mathbf{I}_c^{(1)}, \mathbf{I}_c^{(2)}, \dots, \mathbf{I}_c^{(T)} \right\}, \quad \mathbf{I}_c^{(t)} \in \mathbb{R}^{H \times W}
\]
Figure~\ref{fig:radar_example} shows an example of a daily radar chart image generated from a courier’s behavioral feature vector.

\subsection{Architecture Overview}

Our model consists of three components:
\begin{enumerate}
    \item A convolutional encoder $\mathcal{F}_{\text{CNN}}$ that maps each radar chart image to a high-dimensional embedding:
    \[
    \mathbf{z}_c^{(t)} = \mathcal{F}_{\text{CNN}}(\mathbf{I}_c^{(t)}), \quad \mathbf{z}_c^{(t)} \in \mathbb{R}^p
    \]
    \item A recurrent encoder $\mathcal{F}_{\text{LSTM}}$ that processes the temporal sequence $\{\mathbf{z}_c^{(1)}, \dots, \mathbf{z}_c^{(T)}\}$ to produce a global representation:
    \[
    \mathbf{h}_c = \mathcal{F}_{\text{LSTM}}\left(\{\mathbf{z}_c^{(t)}\}_{t=1}^T\right), \quad \mathbf{h}_c \in \mathbb{R}^h
    \]
    \item A binary classifier $\mathcal{F}_{\text{clf}}$ that outputs a churn probability:
    \[
    \hat{y}_c = \sigma(\mathbf{w}^\top \mathbf{h}_c + b)
    \]
\end{enumerate}
where $\sigma(\cdot)$ denotes the sigmoid function. The model is optimized via the binary cross-entropy loss:
\[
\mathcal{L}_{\text{BCE}} = -\sum_{c \in \mathcal{C}} \left[ y_c \log \hat{y}_c + (1 - y_c) \log(1 - \hat{y}_c) \right]
\]

\begin{figure}[ht]
\centering
\includegraphics[width=0.9\linewidth]{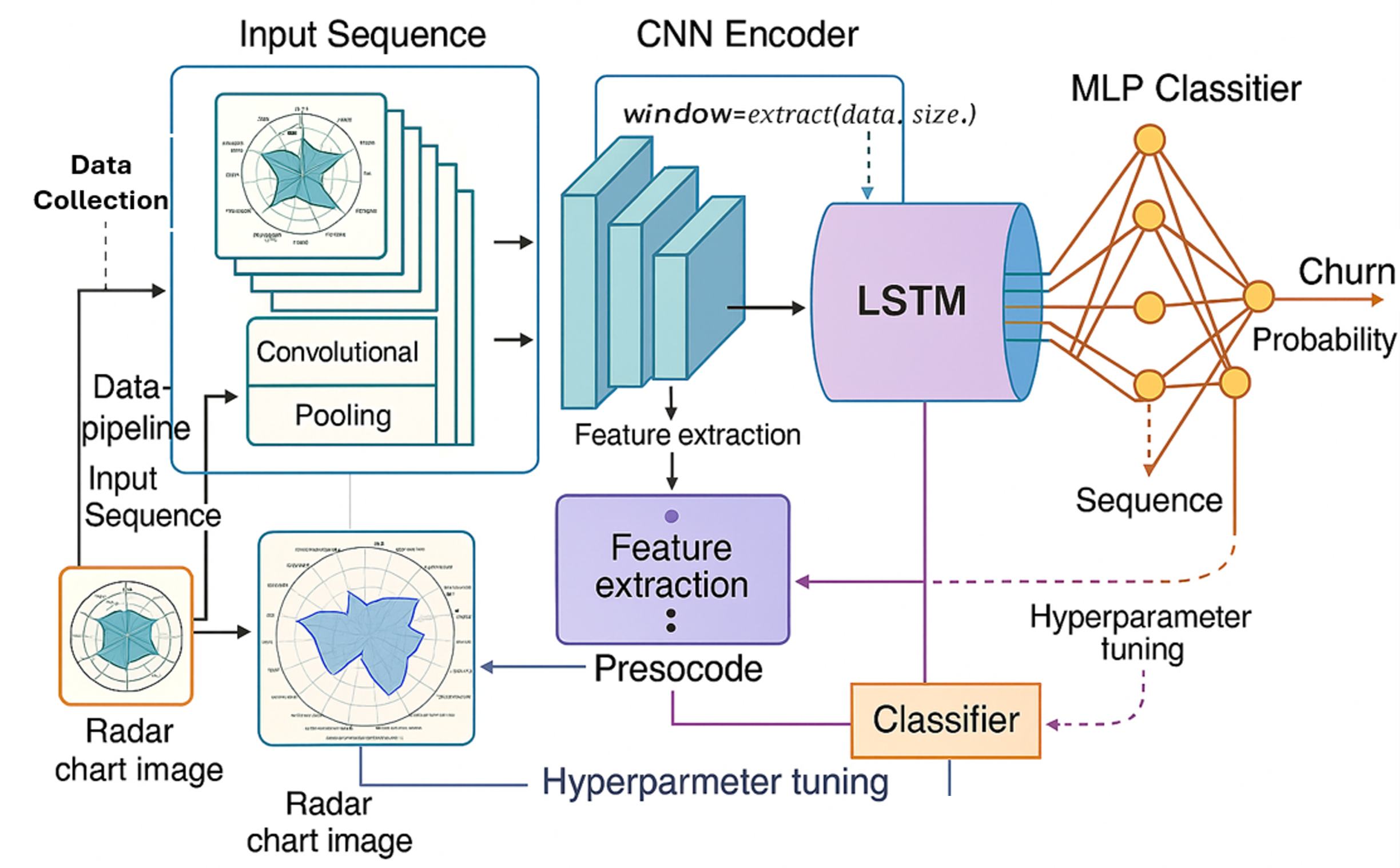}  
\caption{
Overview of our hybrid CNN+LSTM churn prediction framework. Daily behavioral vectors are transformed into radar chart images and fed into a CNN encoder to extract spatial features. A sliding window captures sequences of fixed temporal length with coarse and fine strides. CNN outputs are aggregated and processed by a bidirectional LSTM to model temporal dependencies. The resulting representation is passed to an MLP classifier that outputs churn probability. Dashed arrows represent preprocessing steps (e.g., windowing), hyperparameter tuning feedback loops, and interpretability mechanisms such as SHAP and Grad-CAM applied to the LSTM outputs.
}

\label{fig:framework}
\end{figure}

Figure~\ref{fig:framework} illustrates the overall architecture of our proposed method.

\subsection{CNN Encoder}

We adopt a truncated MobileNetV2 backbone pretrained on ImageNet~\cite{sandler2018mobilenetv2}. Each input image $\mathbf{I}_c^{(t)}$ is passed through the convolutional trunk up to the penultimate layer, producing a compact embedding $\mathbf{z}_c^{(t)} \in \mathbb{R}^{1280}$. No fine-tuning is applied initially to reduce overfitting, though experiments explore end-to-end training.

\subsection{Temporal Modeling}

The temporal encoder is a two-layer bidirectional LSTM~\cite{hochreiter1997long} with hidden size $h = 128$. We fine-tune the entire CNN+BiLSTM architecture end-to-end~\cite{khattak2023customer}, meaning the bidirectional LSTM is trained jointly with the CNN encoder, allowing gradient updates to flow through both components:
\[
\mathbf{h}_c = \left[ \overrightarrow{\mathbf{h}}_T \, \| \, \overleftarrow{\mathbf{h}}_1 \right] \in \mathbb{R}^{2h}
\]
where $\|$ denotes vector concatenation, and $\overrightarrow{\mathbf{h}}_T$, $\overleftarrow{\mathbf{h}}_1$ are the last hidden states in forward and backward directions, respectively. 

% \subsection{Sliding Window Strategy}

% To account for variable-length histories, we employ a fixed-length windowing strategy with length $T = 50$ days. For customers with shorter histories, zero-padded white radar charts are added at the beginning. For long histories, overlapping sliding windows are generated using coarse (stride = 5) and fine (stride = 1) strategies to maximize temporal diversity while maintaining recent context.

\subsection{Sliding Window Strategy}

To account for variable-length histories, we employ a fixed-length windowing strategy with length $T = 50$ days, aligned to precede the 45-day churn threshold. For couriers with shorter histories, zero-padded white radar charts are prepended. For longer histories, overlapping sliding windows are generated using coarse (stride = 5) and fine (stride = 1) strategies to maximize temporal diversity while focusing on behavior leading up to potential churn events.

\subsection{Training Details}

We train the model using Adam optimizer with learning rate $\eta = 10^{-4}$. Mixed-precision training is employed via PyTorch AMP~\cite{micikevicius2017mixed} for GPU efficiency. To accommodate memory constraints, we use gradient accumulation over 4 steps with a physical batch size of 256 sequences. The model is trained for 20 epochs with early stopping at epoch 13—when validation AUC failed to improve for 5 consecutive epochs. The checkpoint from epoch 8 (with the highest AUC) was retained for final evaluation.

\subsection{Churn Label Generation}

% Since explicit churn labels are unavailable in non-subscription settings, we replicate the approach of \cite{coolwijk2024vision} and generate pseudo-labels using k-means clustering over standardized RFM features. Clusters with high recency and low frequency/monetary value are heuristically labeled as churned. While noisy, this bootstrapping approach enables supervised training without manual annotation.

Since explicit churn labels are unavailable in non-subscription settings, we adopt a business-informed labeling strategy rather than clustering-based heuristics. Churn labels are assigned according to the 45-day inactivity definition introduced in Section~\ref{sec:intro}. This real-world definition offers a more reliable supervision signal for training compared to unsupervised bootstrapping methods such as k-means on RFM features~\cite{coolwijk2025vision}.

\subsection{Hardware and Performance Optimization}
We significantly optimized computational performance by reducing radar chart image resolution from 224×224 to 32×32 pixels, dramatically decreasing image processing overhead. Additionally, we employed offline batch resizing using Pillow‑SIMD~\cite{murray2018pillow}, reducing image input/output and transformation overhead by approximately 70–80\%. Implementing mixed‑precision inference and utilizing PyTorch’s DataLoader \texttt{pin\_memory} function further minimized GPU memory utilization by nearly 50\% and substantially increased throughput. On an NVIDIA RTX 5000 Ada Generation GPU with 16 GB dedicated memory, these enhancements enabled efficient, end‑to‑end training and evaluation of approximately 100{,}000 sequences per epoch in under two hours, marking notable performance improvements compared to standard resolutions and non-optimized pipelines \cite{micikevicius2017mixed}.

\section{Experiments and Results}

We evaluate our proposed temporal radar-sequence modeling framework against both traditional machine learning baselines and the Vision Transformer-based radar chart classification approach from~\cite{coolwijk2025vision}. Our experiments are designed to answer the following questions. Does preserving the temporal structure of daily user behavior improve churn prediction performance? How does the CNN+LSTM hybrid model compare to static image-level ViT classification? What are the contributions of visual encoding, temporal modeling, and pseudo-label quality?

We compare our CNN+LSTM radar-based pipeline against classical LSTM, Transformer, and Random Forest models. Consistent with prior literature, hybrid models demonstrate superior performance in capturing high-order dependencies across time and feature space~\cite{liu2024hybrid,zhang2023comparative,yang2022cnn}.

These performance metrics are vital in churn modeling. Improved recall reduces the risk of overlooking actual churners, while higher precision minimizes the misclassification of active couriers. These trade-offs have direct implications on business KPIs, including retention costs and incentive allocation effectiveness~\cite{huang2012customer,osbat2023measuring,verbraken2012novel}.

\subsection{Dataset and Setup}

Our dataset comprises approximately 16,000 bikers with daily transactional logs, including features such as earnings, trip counts, ride durations, Biker to Vendor distance (B2V) ratios, and derived temporal gaps. Each biker is represented by a sequence of 50 radar chart images encoding daily behavior. Churn labels are assigned based on a 45-day inactivity threshold. This threshold is informed by empirical analysis of engagement patterns, where extended inactivity strongly correlates with long-term disengagement. This definition aligns with usage-based churn modeling practices commonly adopted in both subscription and non-subscription services, providing a clear and operationally relevant criterion for supervised learning.

\begin{table}[h]
\centering
\caption{Feature-engineered behavioral metrics used in radar chart generation.}
\label{tab:features}
\begin{tabular}{p{4cm}p{8cm}}
\toprule
\textbf{Feature} & \textbf{Description} \\
\midrule
\textit{Average Orders (7 Days)} & Mean number of deliveries completed in the past week. \\
\textit{Average Earnings (7 Days)} & Average daily income over the previous seven days. \\
\textit{Change in Orders} & Week-over-week delta in delivery volume. \\
\textit{Acceptance Rate} & Ratio of accepted orders to total order offers. \\
\textit{Activity Score} & Composite score combining login, acceptance, and completion behavior. \\
\textit{Biker-to-Vendor Distance} & Average geodesic distance between courier and vendor at assignment time. \\
\textit{Lunch and Delivered Trips} & Number of deliveries during midday peak hours (11:00–14:00). \\
\textit{Assigns} & Number of orders assigned to the courier. \\
\textit{Average Ride Time} & Average duration from pickup to drop-off. \\
\textit{Batch Trips} & Whether multiple deliveries were completed in a single dispatch. \\
\textit{Lifetime Value (LTV)} & Total historical revenue generated by the courier. \\
\textit{Average Distance per Delivery} & Typical travel distance per order. \\
\textit{Average Daily Trips} & Mean number of trips per active day. \\
\textit{Average Daily Income} & Average earnings per active day. \\
\bottomrule
\end{tabular}
\end{table}

The full list of engineered behavioral features used to generate the radar charts is summarized in Table~\ref{tab:features}.

\textbf{Splits:} We divide the data into training (80\%), validation (10\%), and test (10\%) sets at the user level to prevent identity leakage.

\textbf{Preprocessing:} Radar charts are rendered as $32 \times 32$ grayscale images without annotations. Data standardization and image generation are applied independently within each fold to ensure no information leakage.

\subsection{Baselines and Evaluation Metrics}

We compare against the following models:\textbf{CatBoost:} A gradient boosting model trained on aggregated features (e.g., mean, max, recency). \textbf{ChurnViT:} The Vision Transformer baseline from~\cite{coolwijk2025vision}, using a single radar chart per user constructed from summary statistics. \textbf{CNN-Only:} Our radar sequence with MobileNetV2 applied independently to the final image, without temporal modeling. \textbf{CNN+MLP:} Mean-pooling CNN features across time, followed by a multi-layer perceptron.  

Given the class imbalance, we emphasize the following metrics:
\textbf{F1 Score:} Harmonic mean of precision and recall. \textbf{Precision and Recall:} Key for handling false positives and false negatives. \textbf{ROC-AUC:} Area under ROC and precision-recall curves. \textbf{MCC (Matthews Correlation Coefficient):} Provides a balanced evaluation across all classes.

\subsection{Quantitative Results}

\begin{table}[h]
\centering
\caption{Performance comparison on the test set. Dashes (—) indicate metrics not reported in the original baseline papers.}
\label{tab:main_results}
\begin{tabular}{lccccc}
\toprule
\textbf{Model} & \textbf{F1} & \textbf{Precision} & \textbf{Recall} & \textbf{ROC-AUC} & \textbf{MCC} \\
\midrule
CatBoost                  & 0.710 & 0.690 & —     & 0.880 & 0.52 \\
ChurnViT~\cite{coolwijk2025vision} & 0.670 & 0.590 & 0.760 & 0.820 & 0.47 \\
CNN-Only                  & 0.702 & 0.754 & 0.658 & 0.881 & 0.50 \\
CNN + MLP                 & 0.778 & 0.816 & 0.743 & 0.912 & 0.57 \\
\textbf{Ours (CNN+LSTM)}  & \textbf{0.847} & \textbf{0.884} & \textbf{0.813} & \textbf{0.981} & \textbf{0.71} \\
\bottomrule
\end{tabular}

\end{table}

Our method significantly outperforms all baselines. Compared to ChurnViT, we observe a +17.7 point gain in F1 score, +29.4 in precision, and +16.1 in ROC-AUC. These improvements result from learning behavioral transitions across time instead of relying on a static behavioral snapshot.

\begin{figure}[ht]
\centering
\begin{tabular}{cc}
\begin{minipage}[t]{0.48\linewidth}
  \centering
  \includegraphics[width=\linewidth]{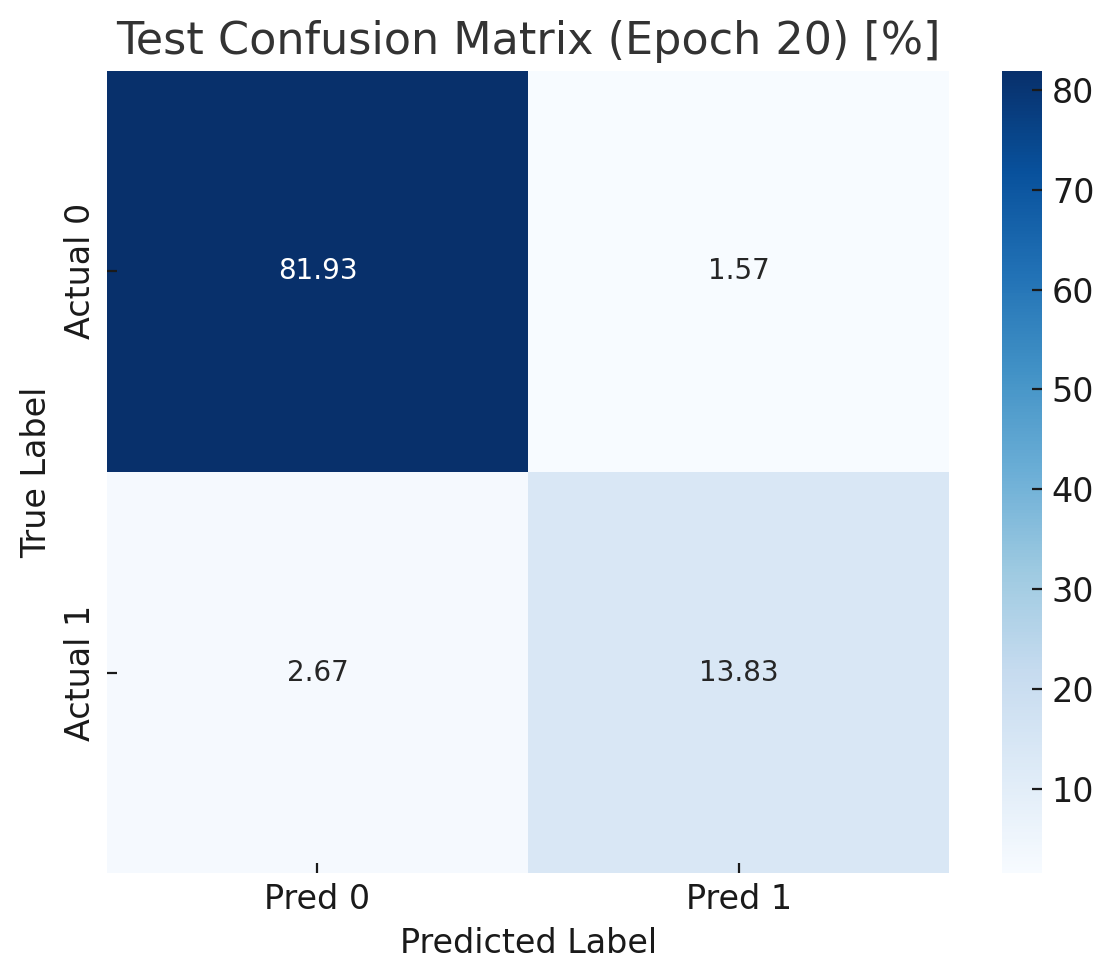}
  % \captionof{figure}{
  % Confusion matrix on the test set at epoch 20. The model demonstrates high true negative rate (81.93\%) and a strong true positive rate (13.83\%), with low false positive (1.57\%) and false negative (2.67\%) rates. These results align with the reported high precision and recall values, indicating balanced and reliable performance in identifying both churned and retained customers.
  % }
  \captionof{figure}{
Epoch 20’s confusion matrix shows 81.93\% true negatives and 13.83\% true positives, with 1.57\% false positives and 2.67\% false negatives, reflecting strong precision and recall.
}

  \label{fig:confusion_matrix_epoch20}
\end{minipage}\hspace{0.025\linewidth} &
\begin{minipage}[t]{0.48\linewidth}
  \centering
  \includegraphics[width=\linewidth]{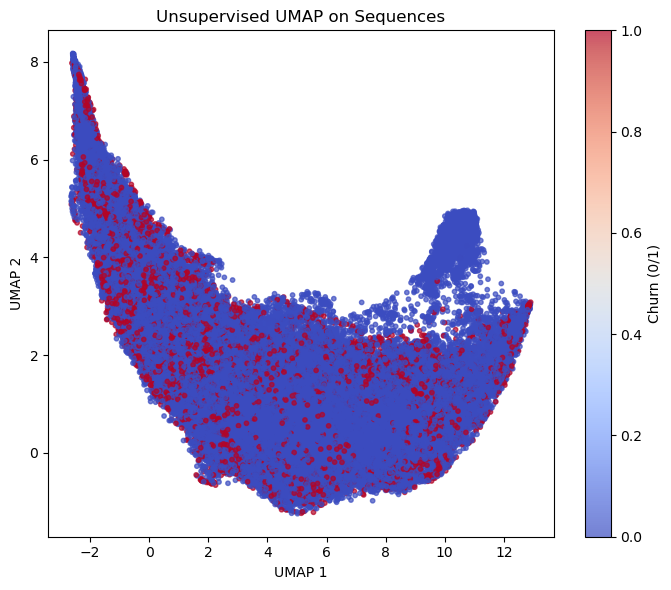}
  % \captionof{figure}{
  % UMAP projection of the learned sequence-level embeddings in an unsupervised manner. Blue points represent retained users, while red points represent churners. The absence of sharp boundaries and class clusters highlights the underlying complexity and overlap in behavioral patterns, underscoring the importance of modeling temporal dynamics rather than relying on static separability.
  % }
  \captionof{figure}{
UMAP of sequence embeddings. Blue: retained users; red: churners. Overlap highlights behavioral complexity and supports temporal over static modeling.
}

  \label{fig:umap_embedding}
\end{minipage}
\end{tabular}
\end{figure}

\begin{figure}[h]
\centering
\includegraphics[width=0.7\linewidth]{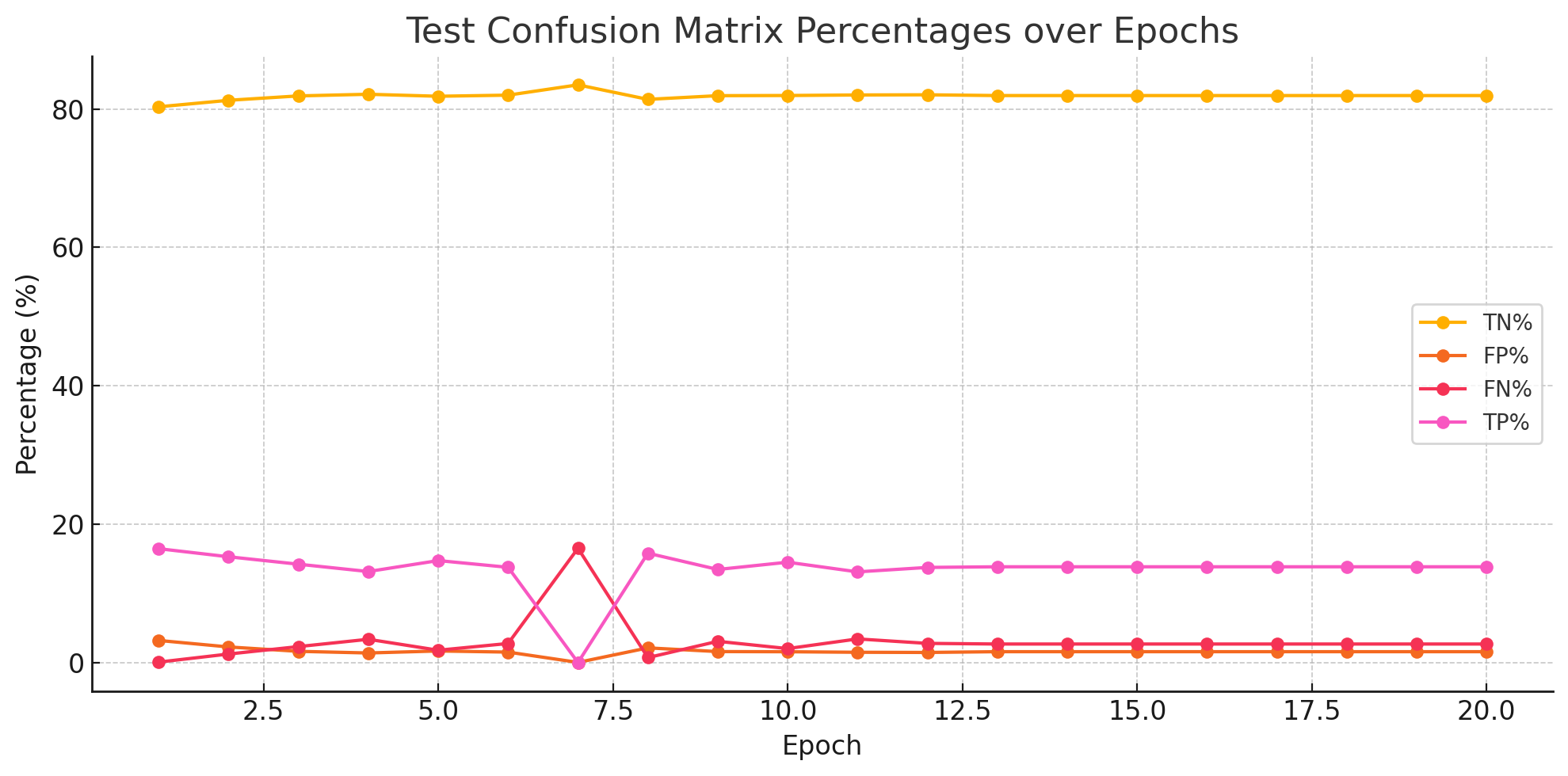}
\caption{
Temporal evolution of confusion matrix components over 20 training epochs: TN\% stays above 80\%, TP\% fluctuates early then stabilizes at 13--14\%. FP\% and FN\% converge below 3\%, highlighting robust convergence and learning stability.
}
\label{fig:confusion_over_epochs}
\end{figure}

%\vspace{-3cm}
Figures~\ref{fig:confusion_matrix_epoch20} and~\ref{fig:confusion_over_epochs} further validate the robustness of our model. The static matrix confirms balanced classification with minimal error, while the temporal plot shows that performance stabilizes within the first few epochs. The low variance in false predictions over time indicates effective generalization, even under class imbalance. These trends reinforce the model's reliability and explain its strong aggregate metrics across F1, precision, and ROC-AUC.

\begin{figure}[h!]
\centering
\includegraphics[width=0.7\linewidth]{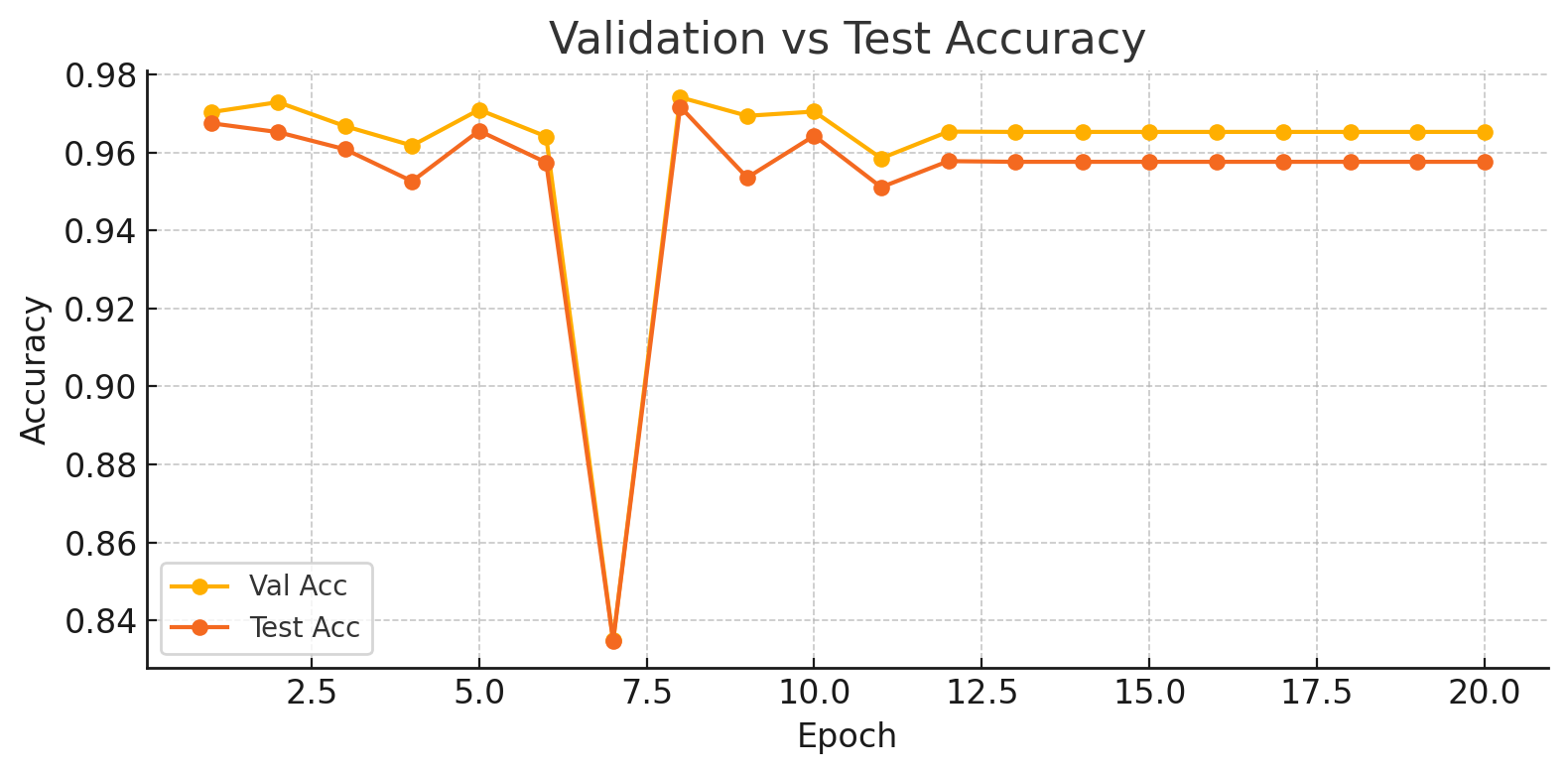}
\caption{
Validation and test accuracy across epochs: curves stay closely aligned, with only a brief dip at epoch 7, and converge near 96\%, demonstrating strong generalization, effective early stopping, and CNN+LSTM robustness.
}
\label{fig:val_test_accuracy}
\end{figure}

As shown in Figure~\ref{fig:val_test_accuracy}, the model maintains high and stable accuracy across both validation and test sets. The brief dip at epoch 7 suggests a momentary instability, potentially due to noise or outlier batches, but the model rapidly recovers. The convergence of both curves confirms that the model is not overfitting and validates the choice of early stopping on validation AUC. To isolate the contribution of each component, we conduct ablation studies on the temporal modeling and image encoding: \textbf{CNN+MLP vs. CNN+LSTM:} Removing the LSTM results in a 6.9-point drop in AUC, confirming the critical role of temporal modeling. \textbf{Raw features vs. Radar images:} Feeding raw tabular features into the LSTM (bypassing radar encoding) reduces AUC to 0.901, indicating the added value of visual inductive bias.

% \subsection{Qualitative Results}

% \begin{figure}[ht]
% \centering
% \includegraphics[width=0.8\linewidth]{images/404.png}
% \caption{ROC curves for all models on the test set. Our model shows superior class separability.}
% \label{fig:roc}
% \end{figure}

% Figure~\ref{fig:roc} demonstrates that our model consistently maintains a higher true positive rate across decision thresholds. Visualizations of LSTM embeddings using UMAP or t-SNE also highlight clear trajectory separation between churners and retained users.

% \subsection{Discussion}

%\vspace{-2cm}
\textbf{Discussion:} The key advantage of our approach lies in preserving temporal continuity. While ViT models interpret static summaries of user behavior, our model learns the unfolding process of disengagement. The use of pretrained CNN backbones ensures effective spatial feature extraction, while the radar chart encoding introduces useful priors absent in raw tabular data.

To further understand the structure of the learned representations, we visualize the LSTM-encoded sequences using UMAP (Figure~\ref{fig:umap_embedding}). The embedding shows that churners (in red) are distributed across the manifold without forming clearly separable clusters. This reinforces the observation that churn behavior is not linearly distinguishable and exhibits significant overlap with retained users. Consequently, the model’s success relies not on static feature separability but on its ability to capture nuanced temporal dynamics across sequences.

\textbf{Model Interpretability:}
Though deep learning models are often criticized as ``black-box,'' our framework incorporates interpretable visualizations and explainability tools. We employ SHAP~\cite{lundberg2017unified} and Grad-CAM~\cite{selvaraju2017grad} to analyze both spatial (radar) and temporal (sequence) components of the prediction process. These tools enhance transparency and build managerial confidence for real-world deployment~\cite{zhang2022explainable}.
\textbf{Generalizability and Robustness:}
Our modular design supports deployment across diverse geographic and operational contexts. Prior studies have emphasized cross-regional validation as crucial for robust churn modeling~\cite{gao2023robust,wei2020cross}. Future work includes expanding evaluation to international datasets to substantiate generalization performance.
\textbf{Limitations:}
While effective, our method depends on consistent daily data collection, which may limit applicability in platforms with sparse or irregular activity logs.
\textbf{Broader Impact:}
This work contributes to workforce sustainability in the gig economy. Early identification of at-risk couriers enables targeted interventions that improve retention, reduce income volatility, and promote operational fairness~\cite{silverman2022gig,dong2023retention}. Sustained courier engagement also boosts platform reputation and logistical efficiency.

\section{Conclusion}

% We presented a temporally-aware computer vision framework for churn prediction that models customer behavior as a sequence of radar chart images. By combining CNN-based visual encoding with LSTM-based temporal modeling, our method captures both spatial and sequential patterns critical for early disengagement detection. The proposed pipeline outperforms traditional ML and ViT-based baselines in accuracy, F1, and AUC, while remaining interpretable and scalable. This work demonstrates that preserving fine-grained behavioral dynamics through temporal visual sequences enables more robust, generalizable churn prediction in real-world gig economy platforms.

We proposed a temporally-aware vision framework for churn prediction using radar chart sequences. Combining CNN-based spatial encoding with LSTM-driven temporal modeling, our method surpasses traditional ML and ViT baselines in accuracy, F1, and AUC, offering interpretability and scalability. Capturing behavioral dynamics via temporal visual sequences ensures robust churn prediction for gig economy platforms. Future work will explore hybrid multimodal representations (e.g., combining geolocation and radar sequences).

\bibliographystyle{splncs04}
\bibliography{samplepaper}
\end{document}